\documentclass{article}

\PassOptionsToPackage{numbers, compress}{natbib}

\usepackage[preprint]{neurips_2026}
\usepackage{amsmath}
\usepackage{amsthm}
\usepackage{amsfonts}
\usepackage{amssymb}
\usepackage{graphicx,psfrag,epsf}
\usepackage{adjustbox}
\usepackage{enumerate}
\usepackage{xcolor}
\usepackage{natbib}
\usepackage{multirow}
\usepackage{multicol}
\usepackage[utf8]{inputenc}
\usepackage{algorithm}
\usepackage[noend]{algpseudocode}
\usepackage{geometry}
\algnewcommand\algorithmicinput{\textbf{Input:}}
\algnewcommand\Input{\item[\algorithmicinput]}
\algnewcommand\algorithmicoutput{\textbf{Output:}}
\algnewcommand\Output{\item[\algorithmicoutput]}
\usepackage{booktabs}
\usepackage{url} 
\usepackage{color,soul}
\usepackage{graphicx}

\DeclareMathOperator*{\argmin}{arg\,min}

\definecolor{azure}{rgb}{0.0, 0.5, 1.0}

\usepackage{xr}

\numberwithin{equation}{section}

\newtheorem{theorem}{Theorem}[section]

\newtheorem{corollary}[theorem]{Corollary}

\newtheorem{proposition}[theorem]{Proposition}

\usepackage[utf8]{inputenc} 
\usepackage[T1]{fontenc}    
\usepackage{hyperref}       
\usepackage{url}            
\usepackage{booktabs}       
\usepackage{amsfonts}       
\usepackage{nicefrac}       
\usepackage{microtype}      
\usepackage{xcolor}         

\title{HALO: Hierarchical Evidence Selection and Reinforcement Learning for Allocating Budgeted Label Optimization from Longitudinal Clinical Records}

\author{%
  Huaiyuan Ying\thanks{These authors contributed equally to this work.} \\
  Department of Statistics and Data Science \\
  Tsinghua University \\
  Beijing, China \\
  \texttt{yinghy22@mails.tsinghua.edu.cn}
  \And
  Feiqing Huang\footnotemark[1] \\
  Department of Biostatistics \\
  Harvard T.H. Chan School of Public Health \\
  Boston, MA, USA \\
  \texttt{fqhuang@hsph.harvard.edu}
  \AND
  Ryumei Nakada \\
  Department of Biostatistics \\
  Harvard T.H. Chan School of Public Health \\
  Boston, MA, USA \\
  \texttt{ryumei\_nakada@hms.harvard.edu}
  \And
  Katherine P. Liao \\
  Brigham and Women's Hospital \\
  Harvard Medical School \\
  Boston, MA, USA \\
  \texttt{kliao@bwh.harvard.edu}
  \AND
  Tianxi Cai \\
  Harvard T.H. Chan School of Public Health \\
  Harvard Medical School \\
  Boston, MA, USA \\
  \texttt{tcai@hsph.harvard.edu}
}

\begin{document}

\maketitle

\makeatletter
\renewcommand{\@notice}{}
\makeatother

\begin{abstract}
Learning from extensive longitudinal clinical records--spanning repeated encounters, evolving conditions, and heterogeneous documentation--under a finite labeling budget requires solving two coupled problems: deciding what evidence to keep and deciding which patients deserve higher-fidelity labels. Prior work on active learning and multi-fidelity label acquisition mainly addresses the second problem, while treating the evidence presented to the acquisition policy as fixed. We present \textbf{HALO}, a \textbf{H}ierarchical Evidence Selection and Reinforcement Learning pipeline for \textbf{A}llocating Budgeted \textbf{L}abel \textbf{O}ptimization from Longitudinal Clinical Records. This framework jointly learns what to read and what label source to acquire under budget. HALO first selects informative evidence from longitudinal records and then uses a budget-aware policy to allocate free-, silver-, and gold-fidelity label sources. For frozen policies, our theory characterizes asymmetric acquired-label channels and gives a budget-constrained finite-sample gold-risk bound for a common downstream logistic model. In a heart-failure case study, HALO attained the highest MedAlign acquired-label F1 at six of eight target budgets. A downstream classifier trained on HALO-acquired labels attained the highest external MGB F1 at budgets 15--60 and remained within 0.01 of the best comparator at budgets 75--120.
\end{abstract}

\section{Introduction}

Longitudinal electronic health records (EHRs) capture rich, evolving clinical information across repeated patient encounters, offering valuable opportunities for robust phenotyping, disease progression modeling, and generating real-world evidence beyond single-encounter snapshots \citep{hripcsak2013next,rajkomar2018scalable,beam2018big}. However, realizing this potential faces a key challenge---EHR data can be noisy and fragmented, incomplete, delayed, or too coarse to capture precise disease status or subtypes \citep{gianfrancesco2018potential,tayefi2021challenges}. Moreover, the longitudinal scale of clinical narratives makes the rich contextual information computationally expensive to process using traditional natural language processing (NLP) methods \citep{zeng2019rich,brender2025clinical}.

The rapid progress of large language models (LLMs) has resulted in powerful tools for extracting clinically meaningful attributes from narrative text \citep{yang2025clines,ying2025geniegenerativenoteinformation}. However, LLM-based pipelines incur substantial costs, both computational and financial, making it impractical to apply them uniformly across entire patient cohorts \citep{Klang_2024}. 
This challenge is compounded by variability in EHR data quality, including documentation artifacts, coding errors, and inconsistencies across notes and systems \citep{lewis2023electronic,shahbodaghi2024documentation}. In addition, obtaining high-quality labels from sparse and dispersed clinically relevant information typically requires expert review or algorithmic validation, making scalable labeling unaffordable \cite{wen2024latte}. A practical learning system must therefore solve two coupled problems to unlock the potential of longitudinal EHRs: (1) deciding what evidence to retain from massive longitudinal narratives—determining which notes, segments, and clinical concepts are most informative for the labeling task; and (2) where to spend limited labeling budget across different fidelity levels. 

A large body of prior work studies how to allocate a limited labeling budget when labels differ in both fidelity and cost. In active learning and multi-fidelity acquisition, the learner seeks to spend a finite budget on queries or labelers that most improve downstream utility, for example, by choosing between weaker and stronger label sources \citep{zhang2015active}, handling imperfect annotators \citep{yan2016active}, or acquiring labels at multiple fidelities under explicit budget constraints \citep{li2022batch}. Related weak-supervision methods ask how to leverage cheap, noisy labels in combination with expensive, high-quality labels \citep{ratner2017snorkel}. However, these methods largely assume that the evidence presented to the label-acquisition policy is already clear. That assumption is problematic for longitudinal EHR, where relevant evidence is often buried in heterogeneous notes and note segments.

In the clinical domain, RELEAP \citep{yang2026releap} is a recent example of using reinforcement learning to improve label efficiency under a finite budget. Nonetheless, it operates on fixed patient-level structured EHR features, such as diagnosis, medication, and laboratory codes aggregated over time, rather than on raw longitudinal clinical notes. It is naturally suited to deciding which patients should receive higher-fidelity labels once a patient-level representation is already available. At the same time, it does not explicitly address the upstream question of what evidence should be retained from longitudinal text to construct that representation in the first place. This distinction matters in modern clinical pipelines because much of the evidence relevant to complex labels is dispersed across narrative notes, and processing all notes uniformly with LLM-based tools can be prohibitively expensive. Prior work on hierarchical document modeling shows that exploiting multi-level structure can improve long-document representation learning \citep{yang2016hierarchical}, and recent clinical studies likewise demonstrate the value of modeling note-level and segment-level structure \citep{kim2023clinical}. However, these approaches typically assume that the full record is already available for modeling and that training labels are given. They do not decide what to retain from the record in order to spend limited labeling budget more effectively.

\begin{figure}[t]
    \centering
    \includegraphics[width=\textwidth]{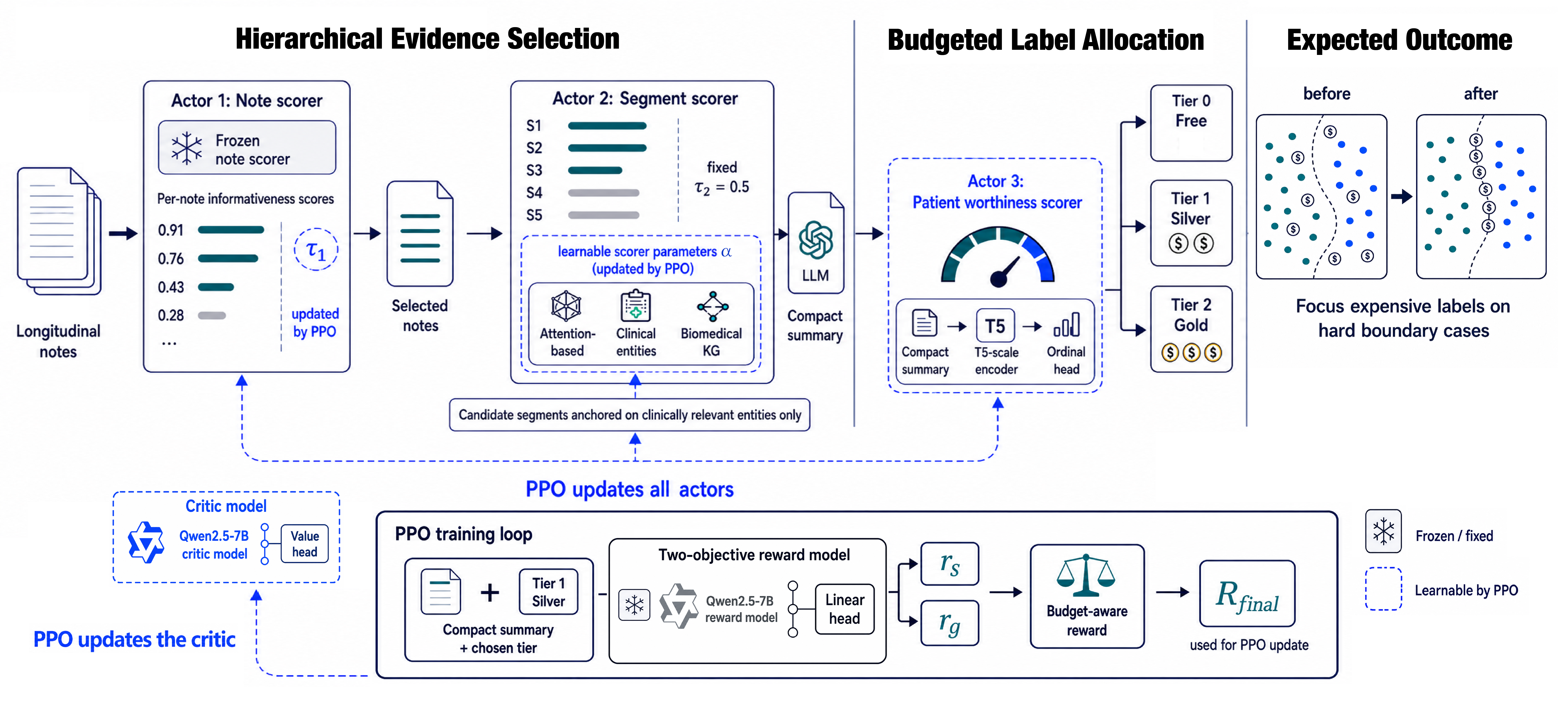}
  \caption{\textbf{Overview of HALO.}
\textbf{Hierarchical evidence selection} first retains informative notes using a frozen note scorer and a fixed threshold $\tau_1$. A trainable segment-informativeness
model with parameters $\xi$ then samples segment-retention decisions during PPO training; at inference, segments are
retained when their predicted retention probabilities exceed $\tau_2=0.5$. The retained evidence is summarized into a compact patient representation.
  \textbf{Budgeted label allocation} feeds the compact summary to a PPO-updated patient-worthiness scorer, implemented with a T5-scale encoder and ordinal head, which routes each patient to free-, silver-, or gold-fidelity label tiers under a finite labeling budget.
  The \textbf{design objective} is to reserve expensive labels for patients for whom the policy predicts higher fidelity to be most valuable.} The bottom panel shows the PPO training loop, where the reward model produces the budget-aware reward used to jointly update the learnable actors and critic.
    \label{fig:main_pipeline}
\end{figure}

In this work, we treat learning from longitudinal records under budget as a coupled decision problem: the learner must decide what evidence to retain and what label source to acquire. We address this problem with \textbf{HALO}, a reinforcement learning framework that jointly performs hierarchical evidence selection and budgeted label allocation. The overall architecture of our approach, spanning from raw clinical notes to budget-aware label assignment, is illustrated in Figure \ref{fig:main_pipeline}. HALO first filters notes and segments to construct a compact patient summary, then allocates limited labeling budget across label sources with different costs and reliability. Throughout the paper, we use \emph{label source} to refer to the annotator or model selected for a patient, \emph{label fidelity} to refer to its free/silver/gold level, and \emph{acquired label} to refer to the label returned by the selected source. We use \emph{downstream} only for the final classifier trained on acquired labels and evaluated on an external cohort.

Our theory characterizes the label channels induced by frozen policies and connects budget-constrained, feature-weighted label corruption to downstream gold risk. Empirically, HALO attained the highest acquired-label F1 on the held-out MedAlign cohort at six of eight target budgets and the highest downstream F1 on the external MGB cohort at budgets 15--60, compared with seven acquisition baselines.

Our contributions are summarized as follows:
\begin{itemize}
    \item We address a key limitation in prior work that treats input evidence as fixed by proposing \textbf{HALO}, a \textbf{H}ierarchical Evidence Selection and Reinforcement Learning pipeline for \textbf{A}llocating Budgeted \textbf{L}abel \textbf{O}ptimization from Longitudinal Clinical Records. It jointly optimizes the two tasks under a finite labeling budget.

    \item We introduce a hierarchical evidence selection mechanism that filters notes and segments to construct compact task-relevant patient summaries, enabling efficient allocation of labeling resources across free-, silver-, and gold-fidelity label sources. 

    \item Theoretically, we characterize asymmetric acquired-label errors and derive a budget-constrained finite-sample gold-risk bound for a well-specified downstream logistic model with nondegenerate features.
\end{itemize}

\section{Methods}

The proposed \textbf{HALO} pipeline first performs hierarchical evidence selection before label allocation, and it is then trained via actor-critic style Proximal Policy Optimization (PPO) \citep{schulman2017proximal}. 
In practice, HALO is applied after an initial cohort-screening step: patients who are clearly irrelevant to the target condition can be filtered out or assigned low-fidelity labels directly, so the budgeted pipeline focuses on potentially relevant cases.

Obtaining the true label of patients requires expert chart review, which is expensive on a large scale. We therefore operate under a finite labeling budget that limits the total label-acquisition cost, and for exposition consider three label-fidelity levels $f,s,g$ which also differ in unit cost, corresponding to free-, silver-, and gold-fidelity label sources as specified in Section \ref{sec:dataset}. The objective of the framework is to maximize label utility as reflected in acquired-label accuracy and the training signal they provide to a post-acquisition predictor. The same framework can accommodate any finite ordered set of label sources with source-specific costs and reliabilities. 

\subsection{Hierarchical Evidence Selection}

The evidence-selection stage follows a coarse-to-fine design. 
The objective of hierarchical evidence selection is to preserve information useful for reliable annotation and budget-aware label allocation while reducing irrelevant documentation. 
For patient \(i\), HALO processes the longitudinal record through
\[
X_i^{(0)} \rightarrow X_i^{(1)} \rightarrow X_i^{(2)} \rightarrow a_i,
\]
where \(X_i^{(0)}\) is the full longitudinal record, \(X_i^{(1)}\) is the record retained after note-level KOMAP filtering, \(X_i^{(2)}\) is the compact patient summary after segment selection and summarization, and \(a_i\in\{f,s,g\}\) is the final free-, silver-, or gold-label-source assignment.

At the note level, we use KOMAP~\citep{xiong2023knowledge}
to compute phenotype-specific informativeness scores from
note-level summary statistics. Notes whose scores exceed
the threshold $\tau_1$ are retained, producing $X_i^{(1)}$.
KOMAP and $\tau_1$ remain fixed during PPO training.

At the segment level, a trainable informativeness model
with parameters $\xi$ scores candidate segments within
$X_i^{(1)}$. We use NILE~\citep{yu2013nile} to extract
Concept Unique Identifiers (CUIs) and obtain their
embeddings. A self-attention layer captures relationships
among concepts within each segment, and a cross-attention
layer uses the target disease PheCode as its query.
For candidate segment $j$ of patient $i$, the model produces
a retention probability
$$
p_{ij}=p_\xi(j\mid X_i^{(1)}).
$$
During PPO training, we sample segment-retention indicators
$$
z_{ij}\sim\operatorname{Bernoulli}(p_{ij}),
\qquad
z_i=(z_{i1},\ldots,z_{im_i}),
$$
where $m_i$ is the number of candidate segments and
$z_{ij}=1$ indicates retention. Conditional on the input,
the segment indicators are sampled independently:
$$
\pi_\xi(z_i\mid X_i^{(1)})
=
\prod_{j=1}^{m_i}
p_{ij}^{z_{ij}}(1-p_{ij})^{1-z_{ij}}.
$$
At inference time, we retain segments deterministically
when $p_{ij}>\tau_2$, with $\tau_2=0.5$.
The retained segments are passed to the fixed summarizer
to produce the compact patient representation $X_i^{(2)}$.
This summary is the interface between evidence selection and budgeted label acquisition: it preserves the salient clinical evidence while sharply reducing the amount of text that the label-allocation policy and higher-cost annotators must inspect

\subsection{Budgeted Label Allocation}

Given $X_i^{(2)}$, the second stage decides whether patient $i$ should receive a free, silver, or gold label under a total labeling budget $\mathcal{C}$. We denote the label-source decision by $a_i \in \{f,s,g\}$ and the corresponding label cost by $c(a_i)$. 
We model this decision using a trainable patient-level worthiness scorer, which is a fine-tuned T5-based INSTRUCTOR model with ordinal head \citep{2020t5,su2023one}. It estimates the labeling difficulty of each patient based on the compact summary. Patients with simpler or more confident evidence are assigned lower-cost labels, while more complex cases are routed to higher-fidelity label sources. The overall objective of allocation is to maximize predictive utility while satisfying the cohort-level budget constraint $\sum_{i=1}^n c(a_i) \le \mathcal{C}$. More concrete formulas are provided in Appendix \ref{appendix:method_details}. 

\subsection{Policy Learning}

\paragraph{Supervised initialization}
We initialize the trainable actors using supervised targets before reinforcement learning. The segment-level informativeness scorer is pretrained on segment informativeness labels, and the patient-level worthiness model is initialized using assignment targets indicating the relative value of free-, silver-, and gold-fidelity labels. KOMAP remains fixed. The reward model is trained separately using utility targets representing improvements from free to silver and from silver to gold labels.

\paragraph{Acquisition policy and trajectory}
Let $\xi$ denote the segment-informativeness model parameters
introduced in Section~2.1, and let $\theta$ denote the
label-allocation parameters. The joint actor parameters are $\omega=(\xi,\theta)$. For patient $i$, the segment-selection action $z_i$ is the binary retention mask sampled from $\pi_\xi(\cdot\mid X_i^{(1)})$. A trajectory sequentially processes patient batches $B_1,\ldots,B_T$. At step $t$, the state before segment selection is
$$
s_t
=\left(\{X_i^{(1)}\}_{i\in B_t},b_t\right),
$$
where $b_t$ is the remaining acquisition budget. For each patient $i\in B_t$, the segment-selection actor samples
$$
z_i\sim\pi_\xi(\cdot\mid X_i^{(1)}).
$$
The retained evidence is processed by the fixed summarizer to produce $X_i^{(2)}$. The allocation actor then samples
$$
a_i\sim\pi_\theta(\cdot\mid X_i^{(2)},b_t),
\qquad a_i\in\{f,s,g\}.
$$
Thus, allocation is conditioned on the representation produced by the preceding evidence-selection decision. The joint batch action is $u_t=\{(z_i,a_i):i\in B_t\}$. Its acquisition cost and the subsequent budget are
$$
C_t=\sum_{i\in B_t}c(a_i),
\qquad
b_{t+1}=b_t-C_t.
$$
The trajectory records the state, sampled action, reward, and next state at each acquisition step. Fixed processing components introduce no separate action-probability terms.

\paragraph{Reward design}
Repeatedly evaluating post-acquisition predictive utility would require costly label acquisition and downstream model training. We therefore use a learned reward model to estimate acquisition utility. For patient $i$, let $\hat q_{i,s}$ and $\hat q_{i,g}$ denote the predicted incremental utility from free to silver and from silver to gold labels, respectively. Denote their within-batch normalized scores by $\bar q_{i,s}$ and $\bar q_{i,g}$. Taking free labels as the reference, the utility assigned to action $a_i$ is
$$
\hat u_i(a_i)
=
\bar q_{i,s}\mathbf{1}\{a_i=s\}
+
(\bar q_{i,s}+\bar q_{i,g})\mathbf{1}\{a_i=g\}.
$$
We aggregate these scores into the batch utility
$$
\widehat U_t
=
\frac{1}{|B_t|}\sum_{i\in B_t}\hat u_i(a_i).
$$
The immediate reward is
\begin{equation}\label{eq:budget_aware_reward}
    R_t
=
\widehat U_t
-\lambda_{\mathrm{budget}}\phi(C_t,b_t),
\end{equation}
where $\lambda_{\mathrm{budget}}$ controls the penalty strength and $\phi$ is the piecewise-linear budget-overrun penalty defined in Appendix~\ref{appendix:method_details}. This penalty discourages overspending but does not impose a hard budget constraint.

\paragraph{PPO optimization}
We jointly optimize both actor components using PPO~\citep{schulman2017proximal}. Unlike the reward model, which scores the current acquisition decision, the value critic $V_\varphi$ estimates the expected cumulative discounted reward from the current state:
$$
G_t=\sum_{k=t}^{T}\gamma^{k-t}R_k,
\qquad
V_\varphi(s_t)\approx
\mathbb{E}_{\pi_\omega}[G_t\mid s_t],
$$
where $\varphi$ denotes the critic parameters and $\gamma$ is the discount factor.

For trajectories collected under $\pi_{\omega_{\mathrm{old}}}$, we calculate generalized advantage estimates~\citep{schulman2015high}:
$$
\delta_t
=
R_t+\gamma(1-d_t)V_{\varphi_{\mathrm{old}}}(s_{t+1})
-V_{\varphi_{\mathrm{old}}}(s_t),
$$
$$
\hat A_t
=
\delta_t
+\gamma\lambda_{\mathrm{GAE}}(1-d_t)\hat A_{t+1}.
$$
Here, $d_t$ indicates a terminal transition, $\lambda_{\mathrm{GAE}}$ controls the weighting of future temporal-difference residuals, and $\hat A_{T+1}=0$ for a completed trajectory. The critic target is
$$
\hat G_t^\lambda
=
\hat A_t+V_{\varphi_{\mathrm{old}}}(s_t).
$$
We standardize the advantages over the collected rollout steps to obtain $\tilde A_t$ for actor training, while retaining the unstandardized advantages when constructing critic targets.

Under the conditional factorization across patients, the joint action probability is
$$
\pi_\omega(u_t\mid s_t)
=
\prod_{i\in B_t}
\pi_\xi(z_i\mid X_i^{(1)})
\pi_\theta(a_i\mid X_i^{(2)},b_t).
$$
The importance ratio compares the probability of the same recorded action under the updated and rollout policies:
$$
\rho_t^{\mathrm{PPO}}(\omega)
=
\frac{\pi_\omega(u_t\mid s_t)}
{\pi_{\omega_{\mathrm{old}}}(u_t\mid s_t)}.
$$
The recorded segment decisions and resulting representations are held fixed when evaluating this ratio. We minimize
$$
\begin{aligned}
\mathcal{L}_{\mathrm{actor}}(\omega)
={}&-\widehat{\mathbb{E}}_t
\left[
\min\left\{
\rho_t^{\mathrm{PPO}}(\omega)\tilde A_t,\,
\operatorname{clip}\left(
\rho_t^{\mathrm{PPO}}(\omega),1-\epsilon,1+\epsilon
\right)\tilde A_t
\right\}
\right]\\
&-\beta_H\widehat{\mathcal{H}}(\omega),
\end{aligned}
$$
where $\epsilon$ controls clipping and $\beta_H$ weights the policy-entropy regularizer $\widehat{\mathcal{H}}$. The leading minus sign converts the maximized PPO objective into a loss to minimize. Both actor components are updated through this joint objective.

The critic minimizes the value-estimation loss
$$
\mathcal{L}_{\mathrm{critic}}(\varphi)
=
c_V\widehat{\mathbb{E}}_t
\left[
\left(
V_\varphi(s_t)
-\operatorname{stopgrad}(\hat G_t^\lambda)
\right)^2
\right],
$$
where $c_V$ weights the value loss. The empirical expectations average over sampled batch decision steps. Rollout-policy parameters, advantage estimates, and critic targets remain fixed during optimization.

\section{Theoretical Analysis}

We analyze how a fixed acquisition policy affects the labels
available for downstream learning. The analysis has two parts:
an oracle comparison of policy-induced label-error channels,
and a finite-sample risk bound for logistic regression trained
on acquired labels with fixed observed features. Both concern
gold-label log loss in the acquisition population. The policies
are held fixed after training; the results do not analyze PPO
optimization dynamics. This perspective complements work on
weak and strong labelers and budgeted multi-fidelity
acquisition~\citep{zhang2015active,yan2016active,li2022batch}.

\paragraph{Fixed policies and acquired labels}
Let $Y_i\in\{0,1\}$ denote the ideal true clinical label,
distinct from the output of any particular acquisition source.
Let $\pi^{(0)}$ and $\pi^{(1)}$ denote fixed acquisition pipelines
without and with hierarchical evidence selection, respectively.
We use $\pi^{(j)}$ as shorthand for a pipeline with its own
fixed actor parameters $\omega_j$. Each pipeline includes
evidence construction and label allocation.

Under policy $j\in\{0,1\}$, patient $i$ has retained
representation $X_i^{(2,j)}$ and receives source assignment
$a_i^{(j)}\in\{f,s,g\}$. Write $Y_i^{(a,j)}$ for the label
returned by source $a$ using the evidence supplied under
policy $j$, and define the acquired label by
$$
\widetilde Y_i^{(j)}
:=
Y_i^{(a_i^{(j)},j)}.
$$
The policy index allows a source's output to change when
the retained evidence changes. If every policy instead
uses the same fixed source outputs, then
$Y_i^{(a,j)}=Y_i^{(a)}$.

For the population comparison, introduce a policy-independent
latent patient state
$$
Z_i=S(X_i^{(0)}).
$$
This state represents clinically relevant information in
the full record and is used only for theoretical conditioning.
The two policies are evaluated under a common patient
population, cohort size, ordering rule, and budget protocol.
The probabilities below refer to a randomly sampled cohort
member and average over the other patients, acquisition
histories, and algorithmic and label-source randomness.

\paragraph{Policy-induced label-error channels}
Define
$$
\eta^*(z)
:=
\mathbb P(Y_i=1\mid Z_i=z),
\qquad
\widetilde\eta_j(z)
:=
\mathbb P_j(\widetilde Y_i^{(j)}=1\mid Z_i=z),
$$
and
$$
\rho_{j,y}(z)
:=
\mathbb P_j\left(
\widetilde Y_i^{(j)}\ne Y_i
\mid Z_i=z,Y_i=y
\right),
\qquad y\in\{0,1\},
$$
where $\mathbb P_j$ denotes the joint law induced by
policy $j$ on the common underlying patient population.
Thus, $\rho_{j,0}$ and $\rho_{j,1}$ describe state-dependent
false-positive and false-negative error, respectively,
permitting asymmetric label noise~\citep{natarajan2013learning}.

\begin{proposition}[Acquired-Label Probability]
\label{prop:signal_strengthening_comparison}
For almost every $z$ and each $j\in\{0,1\}$,
$$
\widetilde\eta_j(z)
=
\eta^*(z)
+
\{1-\eta^*(z)\}\rho_{j,0}(z)
-
\eta^*(z)\rho_{j,1}(z).
$$
Consequently,
$$
\begin{aligned}
\widetilde\eta_1(z)-\widetilde\eta_0(z)
={}&
\{1-\eta^*(z)\}
\{\rho_{1,0}(z)-\rho_{0,0}(z)\}\\
&-
\eta^*(z)
\{\rho_{1,1}(z)-\rho_{0,1}(z)\}.
\end{aligned}
$$
\end{proposition}

This identity follows by conditioning on $Y_i$. It shows
that false-positive and false-negative changes shift the
acquired-label conditional in opposite directions.
Their signed combination, rather than overall label
accuracy alone, determines the change in
$\widetilde\eta_j$.

For a measurable probability function $q$, define the
oracle gold-label population log loss
$$
L(q)
:=
\mathbb E\left[
-Y_i\log q(Z_i)
-
(1-Y_i)\log\{1-q(Z_i)\}
\right].
$$

\begin{corollary}[Oracle Log-Loss Comparison]
\label{cor:signal_strengthening_loss}
Assume
$\eta^*(Z_i),\widetilde\eta_0(Z_i),\widetilde\eta_1(Z_i)
\in(0,1)$ almost surely and that the risks below are finite.
Suppose $\widetilde\eta_1(z)$ lies in the closed interval
with endpoints $\eta^*(z)$ and $\widetilde\eta_0(z)$ for
almost every $z$, and differs from $\widetilde\eta_0(z)$
on a positive-probability set where
$\widetilde\eta_0(z)\ne\eta^*(z)$. Then
$$
\Delta_j
:=
L(\widetilde\eta_j)-L(\eta^*)
=
\mathbb E\left[
\mathrm{KL}\left(
\mathrm{Ber}(\eta^*(Z_i))
\middle\|
\mathrm{Ber}(\widetilde\eta_j(Z_i))
\right)
\right],
\qquad
\Delta_1<\Delta_0.
$$
\end{corollary}

Here, $\mathrm{Ber}(p)$ denotes the Bernoulli distribution
with success probability $p$, and $\mathrm{KL}$ denotes
Kullback--Leibler divergence. The corollary states that
moving the acquired-label conditional toward the true
conditional without crossing it reduces oracle gold-label
log loss. It does not establish that hierarchical selection
produces this movement. It also does not imply the same
risk ordering for a classifier restricted to observed
features: the oracle functions above are evaluated on $Z$,
which the fitted classifier need not observe. We therefore
analyze the observed-feature learner separately below.

\paragraph{Finite-sample learning with fixed observed features}
Let $(V_i,Y_i)_{i=1}^n$ be i.i.d., with
$V_i=v(X_i^{(0)})\in\mathbb R^d$. These features are fixed
before acquisition and shared by both policies. In the
case study, they are note-derived clinical-term counts.
Retained summaries and scoring signals guide acquisition;
they are not additional inputs to this downstream regression.

Throughout, $\|\cdot\|$ denotes the Euclidean norm.
Assume $\|V_i\|\le R<\infty$ almost surely and
$$
\mathbb P(Y_i=1\mid V_i)
=
\sigma(w_*^\top V_i),
\qquad
\lambda_{\min}\left(\mathbb E[V_iV_i^\top]\right)
\ge\nu>0,
$$
where $\sigma(t)=(1+e^{-t})^{-1}$ and
$w_*\in\mathbb R^d$.
For policy $j$, define the signed feature-weighted
label perturbation
$$
d_j
:=
\frac1n\sum_{i=1}^n
\bigl(Y_i-\widetilde Y_i^{(j)}\bigr)V_i.
$$

For the finite-cohort oracle comparison, hold the retained
representations and potential source outputs
$\{Y_i^{(a,j)}\}$ fixed, including any label-source randomness.
For budget $\mathcal C\ge0$, define
$$
D_{n,j}^*(\mathcal C)
:=
\min_{\substack{
b_1,\ldots,b_n\in\{f,s,g\}\\
\sum_{i=1}^n c(b_i)\le\mathcal C
}}
\left\|
\frac1n\sum_{i=1}^n
\bigl(Y_i-Y_i^{(b_i,j)}\bigr)V_i
\right\|,
$$
$$
r_j(\mathcal C)
:=
\|d_j\|-D_{n,j}^*(\mathcal C).
$$
Here, $c(a)\ge0$ and $c(f)=0$, so the feasible set is
nonempty. For each fixed source-output table, the frontier
is nonincreasing in $\mathcal C$, and
$r_j(\mathcal C)\ge0$ for every realized allocation
satisfying $\sum_i c(a_i^{(j)})\le\mathcal C$.
It is an oracle reference requiring true labels and
all potential source outputs; it optimizes source
assignments while holding evidence fixed.
If the source-output table is common across policies,
the frontiers coincide and may be written
$D_n^*(\mathcal C)$.

We analyze the ridge-penalized logistic estimator
$$
\widehat w_j
\in
\argmin_w
\left\{
\frac1n\sum_{i=1}^n
\log\frac{1}{
\sigma\bigl((2\widetilde Y_i^{(j)}-1)w^\top V_i\bigr)
}
+
\frac{1}{2n}\|w\|^2
\right\}.
$$
Its gold-label population risk is
$$
L_Y(w)
:=
\mathbb E\left[
\log\frac{1}{
\sigma\bigl((2Y-1)w^\top V\bigr)
}
\right],
$$
where $(V,Y)$ has the same distribution as $(V_i,Y_i)$.

\begin{proposition}[Budget-Constrained Finite-Sample Risk]
\label{prop:budget_finite_sample}
Under the preceding assumptions, suppose the estimators
under consideration satisfy
$\|\widehat w_j-w_*\|\le B$ almost surely for a deterministic
radius $B>0$. Define
$$
\mu
:=
\nu
\sigma\left(R(\|w_*\|+B)\right)
\left\{
1-\sigma\left(R(\|w_*\|+B)\right)
\right\}
>0.
$$
For every $\delta\in(0,1)$, there is an event of probability
at least $1-\delta$ over the i.i.d. cohort on which,
uniformly over budget-feasible allocations satisfying
the stated localization condition,
$$
\begin{aligned}
L_Y(\widehat w_j)-L_Y(w_*)
\lesssim{}&
\frac{
\{D_{n,j}^*(\mathcal C)+r_j(\mathcal C)\}^2
}{\mu}\\
&+
\frac{R^2\{d+\log(1/\delta)\}}{\mu n}
+
\frac{\|w_*\|^2}{n}.
\end{aligned}
$$
Here, $\lesssim$ suppresses a universal numerical constant.
Acquired labels may be dependent across patients.
\end{proposition}

The first term measures realized feature-weighted label
corruption, decomposed into an oracle frontier and the
allocation's gap above that frontier. The remaining terms
represent finite-sample and regularization contributions.
This decomposition does not establish that PPO achieves
a small allocation gap or satisfies the budget.
The guarantee concerns log loss in the acquisition
population and does not imply an F1 ordering or an
external-domain risk guarantee.
Proofs are provided in
Appendix~\ref{appendix:theoretical_proof}.

\paragraph{Implications for evidence selection and allocation}
Budgeted acquisition should aim to improve the
feature-dependent training signal, and our analysis
characterizes how that signal enters downstream
logistic risk. For fixed downstream features $V_i$,
acquired-label errors induce the gradient perturbation
$d_j$. This motivates jointly designing evidence selection
and label allocation to preserve information that supports
reliable annotation and effective use of higher-fidelity
label sources. 

\section{Experiments}

\subsection{Datasets and Experimental Settings}
\label{sec:dataset}

We evaluated HALO in a heart-failure case study. The target
label was binary: $Y_i=1$ if patient $i$ had ever had heart
failure with preserved ejection fraction (HFpEF), and
$Y_i=0$ otherwise.

\paragraph{Label sources and reference labels}
We considered three label-acquisition sources:
the weakly supervised KOMAP source for Free labels,
Llama~3~8B for Silver labels~\citep{grattafiori2024llama},
and GPT-5 with high reasoning effort for Gold labels.
Throughout the experiments, ``Gold'' denotes the
highest-fidelity acquisition source within this hierarchy,
rather than error-free clinical truth.

Because cohort-wide clinician adjudication was unavailable
for MedAlign, we evaluated acquired labels against a
GPT-5-derived proxy reference. This evaluation measures
agreement with the proxy reference rather than accuracy
against independently established clinical truth.
A clinician separately reviewed all 16 GPT-5-positive
MedAlign patients and 16 randomly selected GPT-5-negative
patients, confirming 13 of the 16 positive annotations
and all 16 negative annotations. Because this review
deliberately sampled by proxy-label status, the overall
agreement of 29/32 should not be interpreted as a
full-cohort annotation accuracy estimate.

\paragraph{Training and evaluation cohorts}
We trained HALO using MIMIC-IV~\citep{johnson2023mimic}.
The PPO training cohort comprised 10,600 patients whose
records contained heart-failure-related descriptions
and longitudinal documentation, with more than one
note from different periods per patient.

The supervised-initialization and reward-model datasets
were also constructed from the MIMIC-IV development
cohort, with patient-level separation from the main
PPO training and evaluation data. Neither these datasets
nor the PPO training cohort overlapped with the MedAlign
acquisition cohort or the external MGB evaluation cohort.
The segment informativeness scorer was initialized using
heart-failure segment annotations, NILE-extracted CUIs,
and GPT-5-mini-derived informativeness labels.
The patient worthiness model was initialized using
patient summaries paired with Free-, Silver-, or
Gold-source assignment targets. The reward model was
trained using Monte Carlo-based patient-level utility
targets representing improvements from Free to Silver
and from Silver to Gold labels.

We deployed the learned acquisition policy on a held-out
MedAlign cohort of 106 patients~\citep{fleming2024medalign},
of whom 16 (15.1\%) had positive GPT-5-derived proxy labels.
For each acquisition strategy and budget setting,
we trained a penalized logistic regression using
the acquired MedAlign labels and patient-level
note-derived clinical term counts.
We evaluated the resulting classifier on an external
Mass General Brigham (MGB) Biobank cohort of 299 patients
against independently curated clinician chart-review
labels. The MGB cohort had an HFpEF prevalence of 44.1\%.

\paragraph{Training and evaluation budgets}
During PPO training, the acquisition budget was set to
10\% of the cost of assigning Gold labels to all patients
included in the training allocation. This setting
determines the total spending allowance rather than
requiring a fixed proportion of Gold assignments.

For evaluation, we used target acquisition budgets of
15, 30, 45, 60, 75, 90, 105, and 120 cost units.
The label-source costs used to generate the reported
results were
$$
c(f)=0,\qquad c(s)=0.6,\qquad c(g)=3.
$$
Thus, one Gold acquisition consumed the same budget as
five Silver acquisitions. These are experimental
relative label-acquisition cost units.

The acquisition budget covers action-dependent
label-generation costs. Evidence extraction,
summarization, and allocation inference are reported
separately as deployment overhead.
In particular, $c(f)=0$ denotes zero additional
label-generation cost for an existing KOMAP label,
rather than zero total computational cost.

\paragraph{Evaluation outcomes and comparators}
We evaluated two distinct outcomes:
(i) acquired-label F1 on MedAlign against the GPT-5-derived
proxy reference, and
(ii) downstream classifier F1 on MGB against clinician
chart-review labels.
The first outcome assesses the acquired labels,
whereas the second assesses their use for training
a classifier evaluated in an external cohort.
Because the cohorts and reference labels differ,
these outcomes should be interpreted separately.

We compared HALO with seven acquisition strategies:
Random assignment, RELEAP~\citep{yang2026releap},
LinTS \citep{agrawal2013thompson}, BADGE~\citep{ash2020badge}, BADGE-Entropy~\citep{ash2020badge}, Core-set~\citep{sener2018coreset}, and DSAL~\citep{zhao2021dsal}.
For RELEAP, we constructed features using the
LLM-powered GENIE entity-extraction
model~\citep{ying2025geniegenerativenoteinformation},
including concept names, counts, and status.

For each acquisition strategy and target budget,
we repeated the experiment ten times and report
the mean F1 across runs for both acquired-label
and downstream classifier performance.
These repetitions use the same patient cohorts;
they characterize algorithmic variability rather
than uncertainty from sampling different cohorts.

\paragraph{Computational resources}
The complete training procedure, including initialization,
reward-model and critic training, and PPO optimization,
required approximately 480 A100 GPU-hours.
These are one-time development costs rather than
costs incurred for each deployment cohort.

A complete frozen-policy deployment run over the
106-patient MedAlign cohort required approximately
30 minutes on three A100 GPUs, corresponding to
1.5 A100 GPU-hours. This measurement includes evidence
scoring, summarization, allocation inference,
model loading, and batching.
Silver- and Gold-label generation costs are accounted
for separately through the acquisition budget.
Additional implementation details and hyperparameters
are provided in Appendix~\ref{appendix:method_details}.

\subsection{Main Results}

\begin{figure*}[ht]
    \centering
    \includegraphics[width=\textwidth]{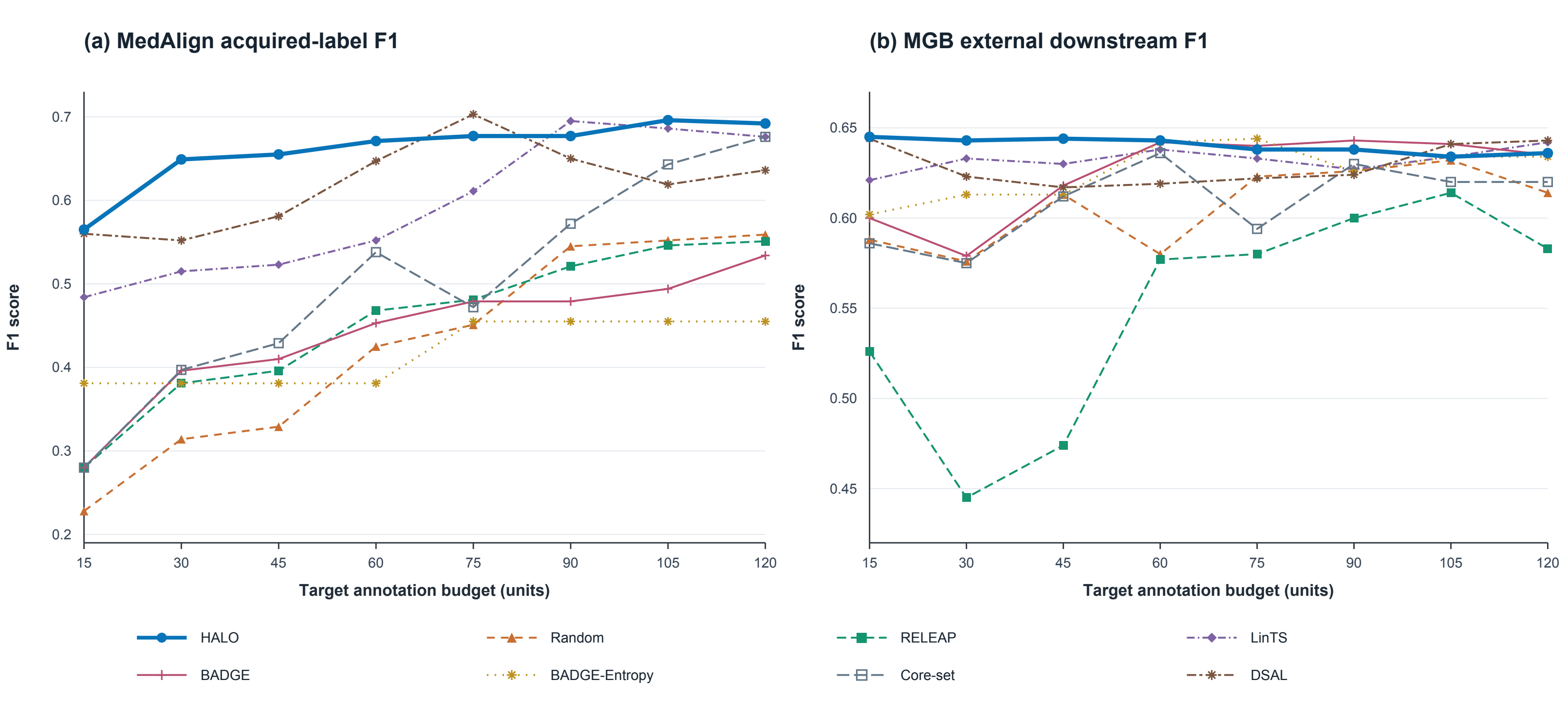}
    \caption{
    \textbf{Performance across target label-acquisition budgets.}
    \textbf{(a) MedAlign acquired-label F1} measures agreement
    between each strategy's acquired labels and the
    GPT-5-derived proxy reference on 106 patients.
    \textbf{(b) MGB external downstream F1} evaluates a
    penalized logistic regression trained on each strategy's
    acquired MedAlign labels against independently curated
    clinician chart-review labels on 299 MGB patients.
    Each plotted value is the mean F1 across ten repeated
    runs. Label-source costs are $c(f)=0$, $c(s)=0.6$,
    and $c(g)=3$.
    Higher values are better in both panels.
    Lines connect the eight evaluated target-budget settings.
    }
    \label{fig:main_results_f1}
\end{figure*}

\paragraph{Acquired-label performance}
Figure~\ref{fig:main_results_f1}(a) reports mean acquired-label
F1 on MedAlign. HALO attained the highest mean F1
at six of the eight target budgets.
Its mean F1 increased from 0.565 at budget 15
to 0.692 at budget 120, reaching 0.696 at budget 105.
DSAL performed best at budget 75
(0.703 versus HALO's 0.677), and LinTS performed best
at budget 90 (0.695 versus HALO's 0.677).
The unweighted average of the eight budget-specific
mean F1 values was 0.660 for HALO and 0.619 for DSAL,
the comparator with the next-highest average.
These comparisons concern agreement with the
GPT-5-derived proxy reference.

\paragraph{External downstream performance}
Figure~\ref{fig:main_results_f1}(b) reports mean downstream
classifier F1 on the clinician-labeled MGB cohort.
The classifier trained on HALO-acquired labels attained
the highest mean F1 at target budgets 15, 30, 45,
and 60, with values of 0.645, 0.643, 0.644, and 0.643,
respectively.
At budgets 75--120, its mean F1 remained within 0.01
of the best-performing comparator at each budget.
These results indicate competitive external downstream
performance, with the clearest advantage in reported
mean F1 at the lower evaluated budgets.

The results distinguish acquired-label agreement from
external predictive performance: improvements in the
former need not produce proportional improvements in
the latter. The comparisons of mean F1 are descriptive
and do not establish statistical significance.
Comparisons at common target budgets also do not,
by themselves, establish equal realized spending
or hard budget compliance.

\section{Limitation}
Our current formulation has several limitations. First, the budget-aware penalty term $\phi(\cdot)$ in the reward function \eqref{eq:budget_aware_reward} discourages overspending but does not impose a strict hard budget, so the policy can exceed the target budget by 3--5\% on average. Second, the framework assumes a fixed taxonomy of three label sources with known costs, and the reward model is trained on a finite set of assignment-outcome pairs, which may limit generalization to new cost structures or acquisition patterns.
In addition, the current design relies on pretrained components (e.g., KOMAP scoring and embedding-based segment selection), which may introduce domain-specific biases and require adaptation across different clinical settings. The downstream finite-sample result assumes a common, well-specified logistic model, a nondegenerate feature second moment, and localization of the fitted coefficient around the gold population minimizer; it does not cover policy-dependent representations or external-domain risk. The GPT-5 annotations used as experimental gold labels are a practical proxy for ideal clinical truth.
Extending the framework to continuous cost spectra, variable per-patient labeling costs, more adaptive evidence selection mechanisms, and more robust reward modeling are natural directions for future work.

\section{Conclusion}

In this work, we presented \textbf{HALO}, a reinforcement-learning framework that jointly performs hierarchical evidence selection and budgeted multi-fidelity label allocation from longitudinal clinical records. For a fixed acquisition policy, our theory characterizes asymmetric acquired-label error and relates a finite-cohort budget frontier and allocation regret to downstream gold risk. In the heart-failure case study, HALO attained the highest MedAlign acquired-label F1 at six of eight target budgets. A classifier trained on HALO-acquired labels attained the highest external MGB F1 at budgets 15--60 and remained within 0.01 of the best comparator at budgets 75--120.

\newpage

\bibliographystyle{unsrt}
\bibliography{budget_constraint}

\newpage

\end{document}